\pdfoutput=1

\documentclass[11pt]{article}

\usepackage{tcolorbox}
\usepackage{xcolor}
\usepackage{soul}
\usepackage{multirow}
\usepackage{acl}

\usepackage{times}
\usepackage{latexsym}
\usepackage{siunitx}

\usepackage[T1]{fontenc}

\usepackage[utf8]{inputenc}

\usepackage{microtype}

\usepackage{inconsolata}

\usepackage{graphicx}

\title{Mathematical Reasoning in Large Language Models:\\Assessing Logical and Arithmetic Errors across Wide Numerical Ranges}

\author{
  Safal Shrestha\textsuperscript{*} \quad Minwu Kim\textsuperscript{*} \quad Keith Ross \\
  New York University Abu Dhabi
}

\begin{document}
\maketitle

\renewcommand{\thefootnote}{\fnsymbol{footnote}}
\footnotetext[1]{Equal contribution; order determined by a coin toss.}
\footnotetext{Correspondence: keithwross@nyu.edu}

\begin{abstract}


Mathematical reasoning in Large Language Models (LLMs) is often evaluated using benchmarks with limited numerical ranges, failing to reflect real-world problem-solving across diverse scales. Furthermore, most existing evaluation methods only compare model outputs to ground-truth answers, obscuring insights into reasoning processes. To address these limitations, we introduce GSM-Ranges, a dataset generator derived from GSM8K that systematically perturbs numerical values in math problems to assess model robustness across varying numerical scales. Additionally, we propose a novel grading methodology that distinguishes between logical and non-logical errors, offering a more precise evaluation of reasoning processes beyond computational accuracy. Our experiments with various models reveal a significant increase in logical error rates—up to 14 percentage points—as numerical complexity rises, demonstrating a general weakness in reasoning with out-of-distribution numerical values. Moreover, while models demonstrate high accuracy on standalone arithmetic tasks, their performance deteriorates substantially when computations are embedded within word problems. These findings provide a comprehensive evaluation of LLMs’ mathematical reasoning capabilities and inform future research directions for improving numerical generalization in language models. \footnotetext{Code and relevant dataset available at \url{https://github.com/minwukim/GSM-Ranges}}

\end{abstract}

\section{Introduction}

\begin{table}[h!]
\centering
\renewcommand{\arraystretch}{1.3} 
\setlength{\tabcolsep}{3pt} 

\begin{tabular}{p{1.0\columnwidth}}
\hline

\textbf{GSM8K} \\

Judy teaches \textbf{\underline{5}} dance classes every day on the weekdays and \textbf{\underline{8}} classes on Saturday. If each class has  \textbf{\underline{15}} students and she charges \$\textbf{\underline{15}} per student, how much money does she make in \textbf{\underline{1}} week? \\ \hline

\textbf{GSM-Ranges} (Level 6 Perturbation)\\

Judy teaches \textbf{\underline{3,124,213}} dance classes every day on the weekdays and \textbf{\underline{7,832,129}} classes on Saturday. If each class has \textbf{\underline{25}} students and she charges \$\textbf{\underline{35}} per student, how much money does she make in \textbf{\underline{1}} week? \\ \hline
\vspace{-0.5em}

\caption{An example of a question generated by \textit{GSM-Ranges} tool, derived from a base problem from the GSM8K dataset.}
\label{tab:gsm_ranges}





\end{tabular}
\end{table}
Mathematical reasoning with Large Language Models (LLMs) has recently been the subject of significant attention \cite{wei2022-CoT,openai2024-gpt4,ahn2024}.
However, the current evaluation methodologies for these systems
exhibit notable limitations. First, existing benchmarks primarily focus on problems with limited numerical ranges \cite{madaan2022-50percent}, leaving a significant gap between the controlled evaluations and real-world settings.
Second, traditional grading approaches typically compare LLMs' final answers directly with the ground truth answers \cite{hong2024caught,shakarian2023eval}, a practice that conflates logical and numerical errors, 
thereby obscuring a deep understanding 
of the LLM's reasoning capabilities. 
This motivates the need for a better evaluation approach that not only handles numbers across wide numerical ranges but also distinguishes between logical and arithmetic errors. 

This paper makes three contributions. First, we introduce publicly available {\em GSM-Ranges}, a tool for generating datasets that are designed to evaluate error rates and error types across diverse numerical ranges. Derived from the GSM8K dataset \cite{cobbe2021-gsm8k}, GSM-Ranges systematically organizes problems into distinct numerical intervals. Specifically, GSM-Ranges applies six distinct levels of perturbations to 
GSM8K questions, replacing existing numbers with random values across 
six distinct scales. 

Second, we introduce a novel grading methodology that distinguishes between logical and non-logical errors. 
We claim that a solution should be deemed logically valid if computational inaccuracies
can be corrected and the revised answer matches the ground truth. Conversely, if the final answer remains incorrect despite eliminating such errors, we infer a fundamental flaw in the reasoning process. 
To automate this assessment, 
our methodology employs GPT-4o model \cite{openai2024-gpt4o} to translate a LLM-generated response into Python code that accurately 
captures the underlying logic. By executing this code, we isolate non-logical errors and compute the corrected final answer, which is then 
compared with the ground truth to assess logical correctness.
We also perform a careful evaluation of our automated grading methodology, 
and confirm its high level of accuracy for distinguishing between the two error types.

Third, using our grading methodology and our GSM-Ranges tool, we analyze
various open-source and proprietary LLM models. This leads to several findings: 
\begin{itemize}
    
    \item Previous works have shown that arithmetic errors become more pronounced for larger numbers \cite{qian2023-arithmeticerror, feng2024numerical}. We find that this trend applies to logical errors as well, with the worst-case logical error rate increasing by up to 14 absolute percentage points as perturbation levels rise. This is surprising since the logical reasoning process required to solve the problems remains unchanged despite the numerical modifications. Nevertheless, we still observe an increase in logical errors as perturbation levels rise, suggesting that the logical reasoning in LLMs tends to exacerbate for out-of-distribution numerical values, compromising their robustness in handling broader numerical scales.
    
    \item Previous studies have demonstrated that LLMs achieve high accuracy on standalone arithmetic tasks with in-distribution numbers (e.g., “36$\times$6=?”) \cite{yang2023gpt, maltoni2024arithmetic, yuan2023well, mirzadeh2024-gsmsymbolic, xie2024llm}. While we confirm these findings, our results further reveal that the accuracy of arithmetic computations significantly deteriorates when the calculations are embedded within word problems (e.g., “36 apples from Jack$\times$6 = ? apples”). 
    
\end{itemize}
By introducing GSM-Ranges and a novel grading methodology, this work aims to provide a more comprehensive evaluation of mathematical reasoning in LLMs. Our approach not only isolates logical and arithmetic errors but also assesses model robustness across a broad range of numerical values. These insights pave the way for future research on improving LLMs’ mathematical reasoning capabilities and developing models that can generalize more effectively across diverse mathematical problem settings.

\section{Related Work}

\textbf{LLM Sensitivity to Perturbations.}
Several prior studies \cite{stolfo2022causal,hooda2024large,jiang2024-peek,guo2024learning} have explored the sensitivity of LLMs to perturbations in input problems, demonstrating significant performance degradation even when the underlying logic remains the same. In the domain of mathematical word problems (MWPs), particularly on the GSM8K benchmark, this degradation has been observed when numbers are slightly changed from the original question. \cite{li2024-gsmplus,mirzadeh2024-gsmsymbolic,shi2023-GSMIC}. However, existing approaches often constrain substituted values to a limited numerical range \cite{stolfo2022causal} or use numbers that remain comparable to the original values, which tend to be small \cite{li2024-gsmplus,mirzadeh2024-gsmsymbolic,madaan2022-50percent}. In this paper, we go beyond the narrow constraints previously studied, providing a comprehensive investigation into how different numerical ranges can impact mathematical abilities in LLMs.\\

\noindent \textbf{Evaluation of Mathematical Correctness.} Prior correctness evaluations (grading) predominantly rely on ground-truth comparisons
\cite{shakarian2023eval,fu2023chain,hong2024caught,frieder2024mathematical}. 
However, such a straightforward approach does not distinguish between logical and non-logical errors, and therefore cannot alone accurately assess an LLM's mathematical reasoning capabilities. 
Previous studies have explored simple prompting strategies for evaluation, finding that while LLMs perform well in generating correct answers to benchmark questions, they struggle to identify and diagnose errors in solutions to those same questions. This difficulty is particularly pronounced for non-logical errors, highlighting a fundamental gap in their problem comprehension \cite{li2024evaluating, zeng2024mrgsm8kmetareasoningbenchmarklarge}. A straightforward alternative involves using external tools, such as calculators, to mitigate non-logical errors. However, this approach requires fine-tuning models to follow a specific format and does not always produce reliable results \cite{schick2023toolformer, cobbe2021-gsm8k}. To address these challenges, we introduce a novel automated grading methodology that eliminates the need for fine-tuning while effectively differentiating between logical and non-logical errors. This enables a more precise assessment of how mathematical reasoning deteriorates under various numerical conditions.




\noindent \textbf{Arithmetic Errors Due to Perturbations.} Past studies have shown that LLMs handle basic arithmetic reasonably well when the arithmetic involves small numbers in standalone queries like \textit{"What is x + y?"}, a skill often linked to memorization \cite{yang2023gpt, maltoni2024arithmetic, yuan2023well, qian2023-arithmeticerror, feng2024numerical}. Performance drops significantly in more complex cases, such as multi-step equations, large numbers, and multiplication \cite{kao2024solving, yuan2023well, yang2023gpt, feng2024numerical, qian2023-arithmeticerror}. Background context can further affect arithmetic reasoning, introducing additional inconsistencies \cite{abedin2025arithmattack}. 
Current research applying perturbations to widely used benchmarks like GSM8K involve basic arithmetic; thus, arithmetic errors are assumed to be minimal. \cite{mirzadeh2024-gsmsymbolic, xie2024llm, anand2024mathify}.  However, a deeper understanding of the nature and frequency of arithmetic errors remains lacking. This study addresses this gap by systematically quantifying arithmetic errors in mathematical word problems and conducting a qualitative analysis to identify underlying error patterns.

\section{GSM-Ranges}

\begin{figure}[t]
  \centering
  \includegraphics[width=0.9\columnwidth]{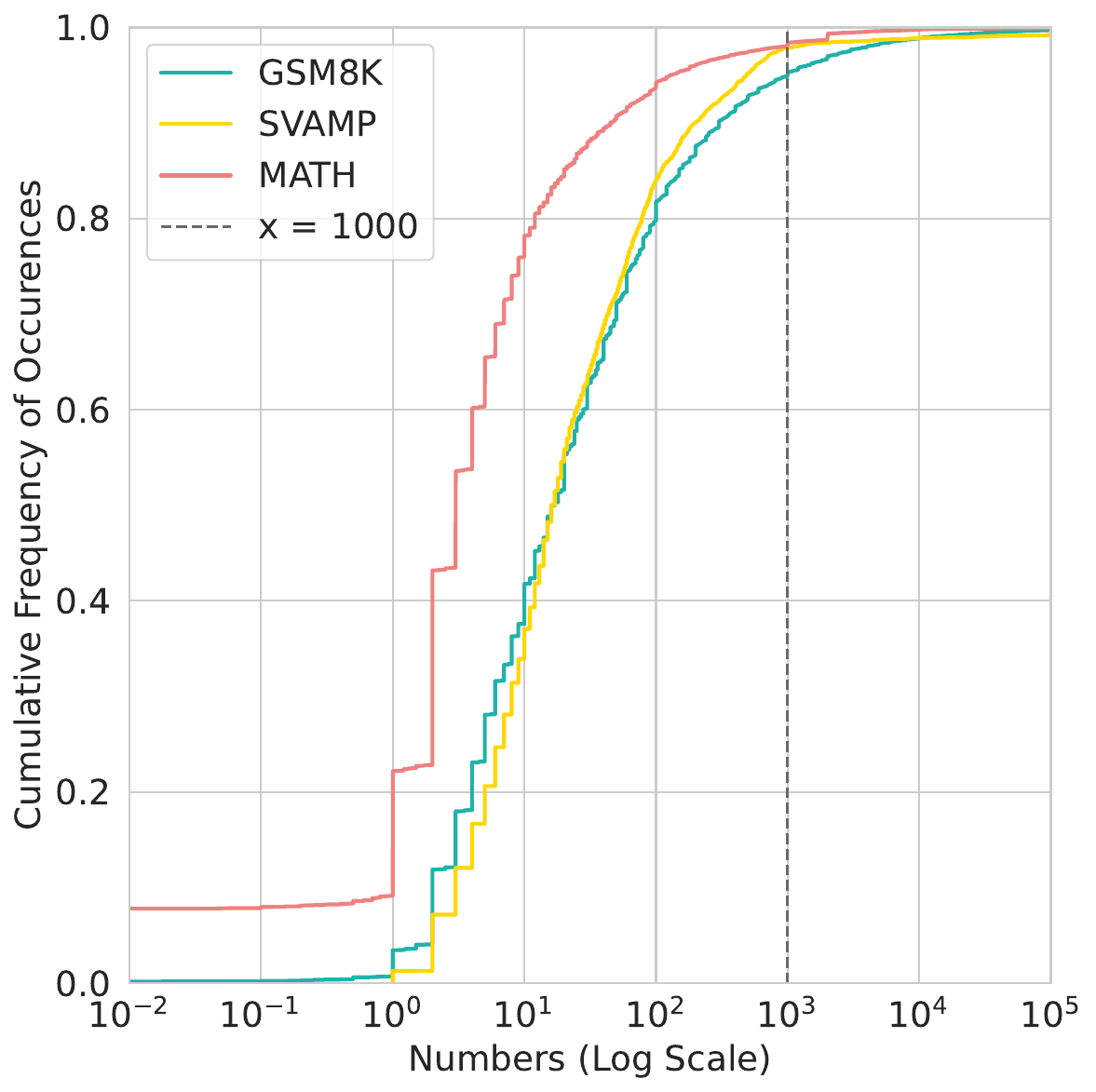}
  \caption{Cumulative frequency distribution of numerical values in questions and ground truth answers. Numbers <1,000 account for 94.9\% (GSM8K), 97.8\% (SVAMP), and 98.0\% (MATH) of the values.}
  \label{cumulativegraph}
\end{figure}

The majority of existing mathematical benchmarks are constrained to relatively limited numerical ranges. For instance, \citet{madaan2022-50percent} reported that single-digit numbers constitute approximately 50\% of the problems in the GSM8K dataset. To further investigate this trend, we analyzed cumulative frequency distribution of numerical values across three widely used benchmarks: GSM8K \cite{cobbe2021-gsm8k}, SVAMP \cite{patel2021-SVAMP}, MATH \cite{hendrycks2021-MATH}. As shown in Figure \ref{cumulativegraph}, our analysis reveals that in all three datasets, numbers below 1,000 (i.e., three digits or fewer) account for 94.9\% of values in GSM8K, 97.8\% in SVAMP, and 98.0\% in MATH. These findings indicate that the mathematical capabilities of current LLMs have primarily been evaluated within a limited numerical range. To assess robustness across wider ranges, we introduce \textit{GSM-Ranges}, a tool for generating datasets which encompass a wider distribution of numerical values.

\subsection{Selecting Base Problems from GSM8K}

GSM-Ranges systematically modifies numerical values in the GSM8K dataset. From the GSM8K test set of 1,319 questions, we exclude those involving non-integer values or division in the ground truth answers, as such cases could potentially create logically incoherent problems (e.g., “Assign 5 people evenly to 2 separate rooms”). After filtering, 100 questions are randomly selected, with all numbers in the questions being single- or double-digit.

\subsection{Perturbation Levels}
We convert the 100 sampled questions into Python templates to systematically adjust the numerical values within the questions across various ranges. Specifically, we apply 6 levels of perturbation: same-digit, 100--1,000, 1,000--10,000, 10,000--100,000, 100,000--1,000,000, and 1,000,000--10,000,00, which we refer to as level 1 to level 6 perturbation, respectively. In same-digit perturbation (level 1), we randomly replace each number in the problem with a randomly chosen number with the same number of digits as the original number, thereby maintaining the similarity of the perturbed problems to the original. Additionally, we ensure that modified numbers are always different from the original values, preventing any duplication of the original problems. In 100-1000 perturbation (level 2), we replace each number in the problem with a number randomly chosen in the range 100 to 1000. Levels 3-6 are done in similar manners with their respective ranges. 

All perturbations ensure non-negative final answers and intermediate values, 
as, similar to the case of fractional values, they can lead to logically incoherent math problems (e.g. “A store sells -7 items in a day”, “Eat 10 apples out of 3 apples”). In addition, to prevent extreme scaling of final answers, we apply scaling selectively in cases involving multiplication. Specifically, when the final answer is derived from a multiplication operation (e.g., $(A+B) \times (C+D-E)$), we scale only one side of the multiplication—either $A$ and $B$ or $C$, $D$ and $E$—while keeping the other side within the original numerical ranges. This approach maintains the final answer within manageable limits while introducing numerical variation in the problem, preventing the inclusion of excessively complex or computationally infeasible arithmetic operations for LLMs. An example of a problem generated with these crafted templates is shown in Table \ref{tab:gsm_ranges}.


\begin{figure}[t]
  \centering
  \includegraphics[width=\columnwidth]{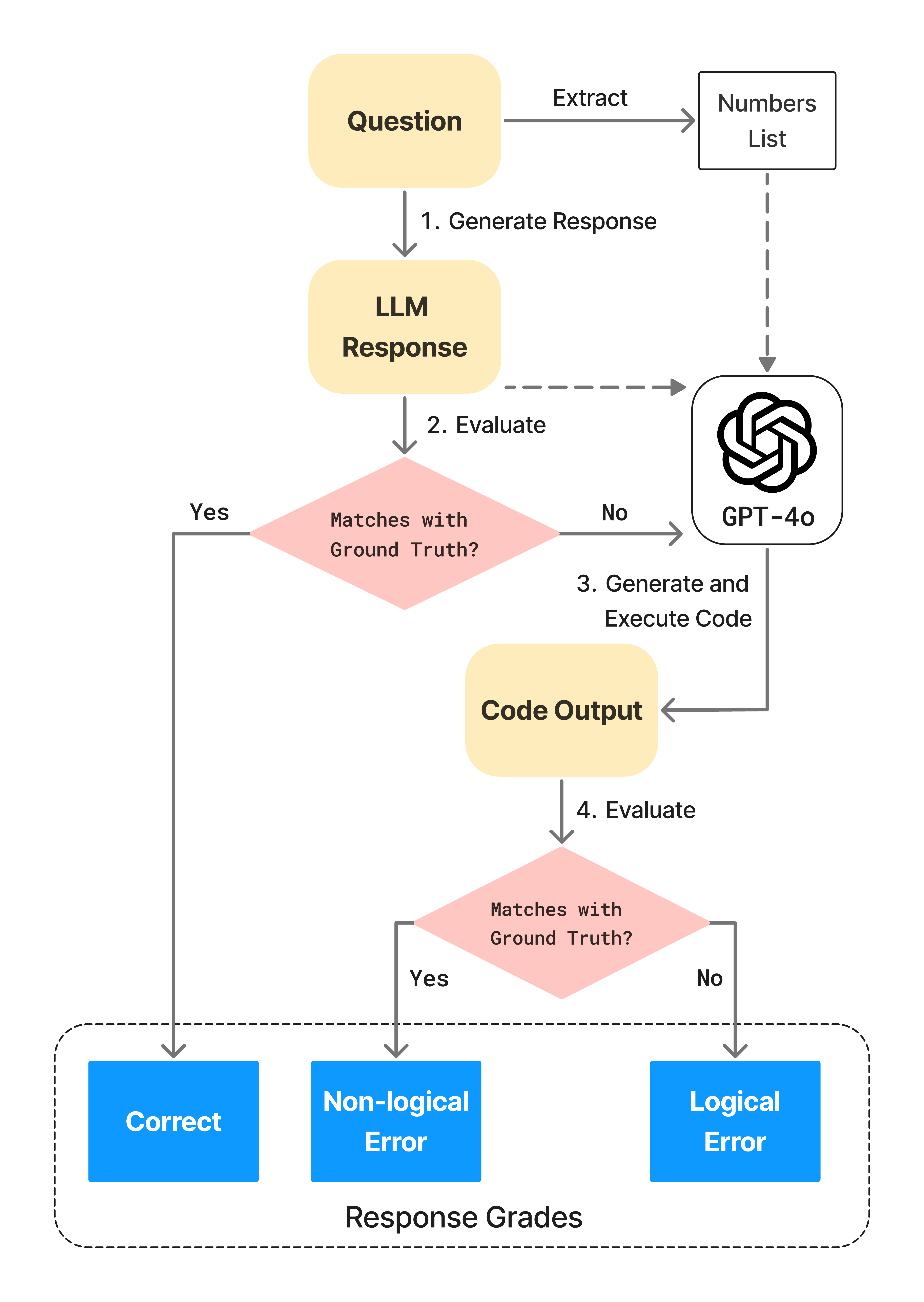}
  \caption{Illustration of grading process for LLM responses using the GPT-4o model, categorizing outputs into three labels: correct, non-logical error, and logical error.}
  \label{grading}
\end{figure}

\begin{figure*}[t]
  \centering
  \includegraphics[width=1.0\textwidth]{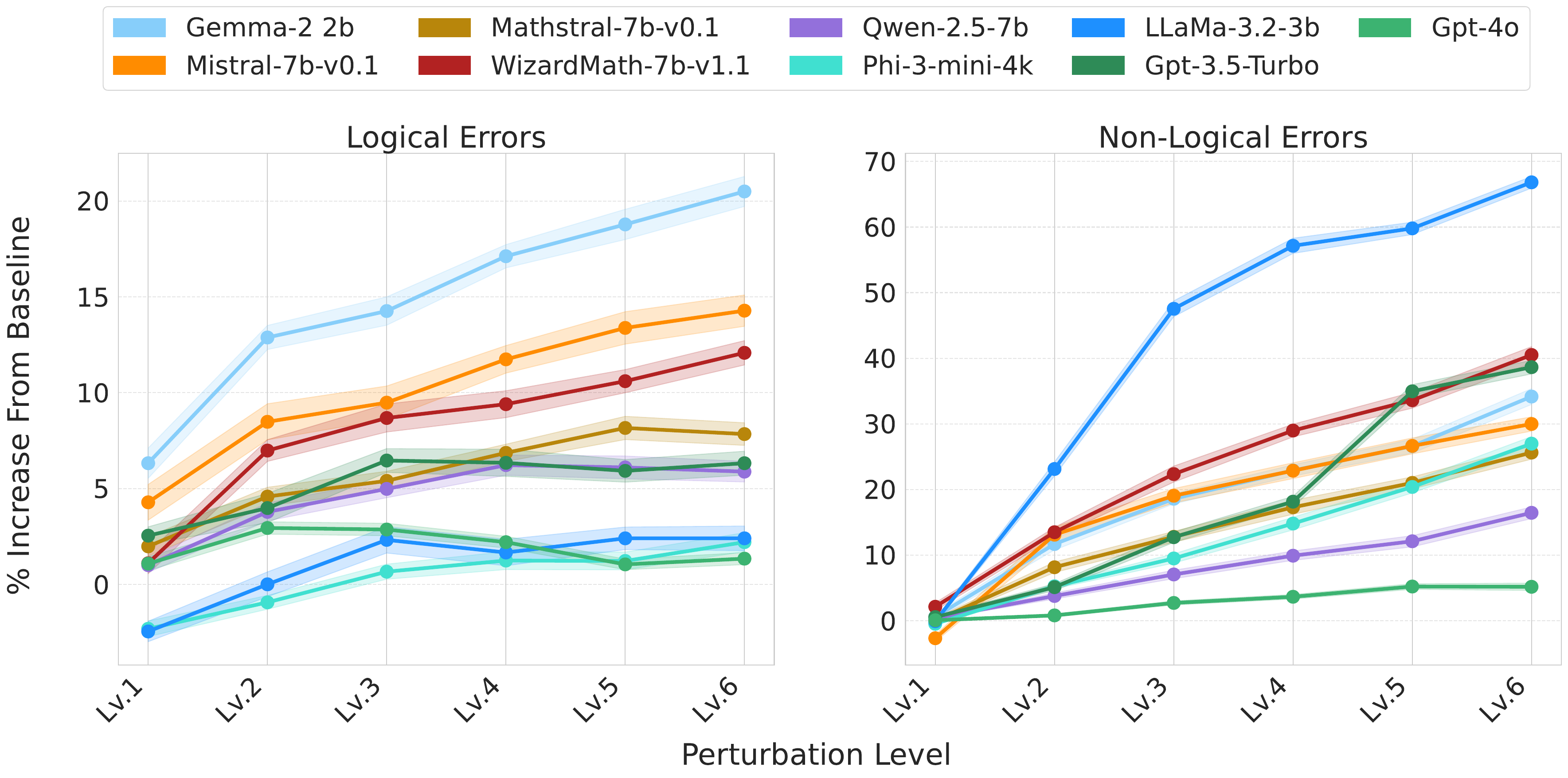}
  \caption{Logical \& non-logical error rates across different perturbation levels. The left panel illustrates the increase in logical errors across the datasets, while the right panel depicts the rise in non-logical errors. Error rates are reported relative to the baseline logical and non-logical error rates on the original GSM8K problems.}
  \label{fig:robust_exp}
\end{figure*}

\begin{figure}[t]
  \includegraphics[width=\columnwidth]{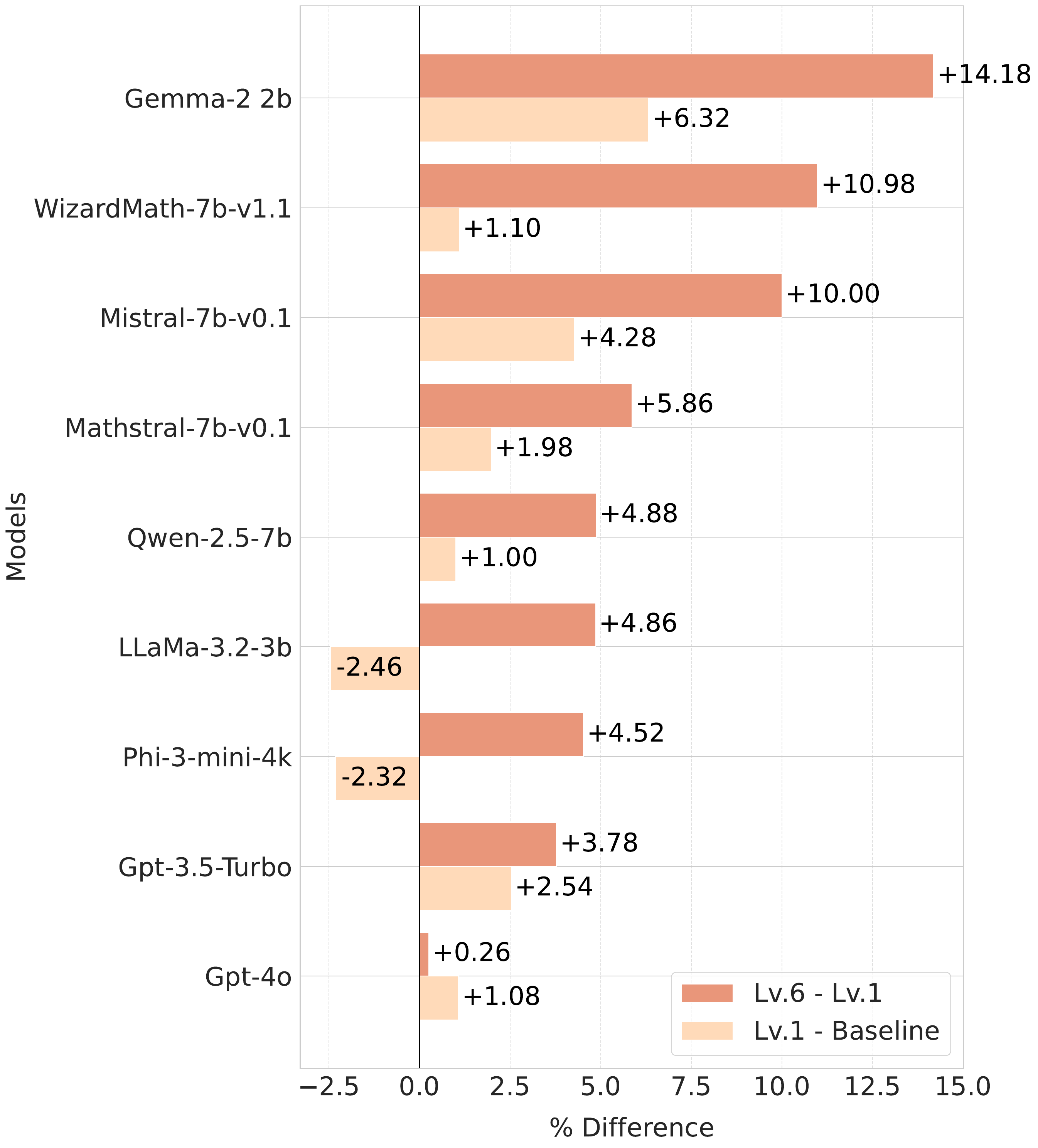}
  \caption{Logical error gaps across perturbation levels. For each model, the top bar represents the percentage point difference in logical errors between Level 6 and Level 1 perturbations, while the bottom bar indicates the percentage point difference between Level 1 and the original GSM8K questions.}
  \label{fig:r6r1gap}
\end{figure}

\section{Grading Methodology}

While modern LLMs perform well on basic arithmetic operations, their accuracy is reported to diminish substantially as numerical magnitudes increase \cite{qian2023-arithmeticerror}. However, conventional evaluation methods, which primarily compare final answers with ground truth values \cite{hong2024caught,shakarian2023eval}, fail to provide a complete picture of LLMs' mathematical understanding as they do not distinguish computational errors from reasoning errors.
To address this deficiency, we  propose a novel and accurate grading methodology which not only determines whether the answer is correct or erroneous, but also categorizes the type of error.


\subsection{Error Definitions}
Our grading methodology first compares the final answer with the ground truth. If the answers match, the response is labeled as correct. Otherwise, there is an error. We define the error types as follows:
\begin{itemize}
    \item \textbf{Non-logical error}: Errors that do not stem from the reasoning process itself, such as arithmetic errors or number-copy errors. The latter refers to inaccurately reproducing problem values (e.g., misrepresenting 1,337,042 as 13,337,042) and is observed in higher-level perturbations in some models.

    \item \textbf{Logical error}: All errors not classified as non-logical, such as but not limited to missing steps, contradictory steps, or operator misuse. It should be noted that responses classified as logical errors may also include non-logical errors.
\end{itemize}

\subsection{Methodology Overview}
To handle the extensive volume of responses, we employ the GPT-4o model \cite{openai2024-gpt4o} to automate the evaluation.
For a response that fails to match the ground truth answer, we have GPT-4o model translate the reasoning in the response into Python code, which is then executed to generate a new answer. The prompt is provided in Appendix \ref{subsec:4o-prompt}.
If the new answer aligns with the ground truth, our methodology classifies the response as containing a non-logical error; otherwise, it is classified as a logical error. A detailed illustration of the grading process is shown in figure \ref{grading}.

\subsection{Number-Copy Errors}
Apart from arithmetic errors, we observe that number-copy errors are non-trivially present in some models under higher levels of perturbations (Appendix \ref{subsec:number copy error}). For instance, in the case of the Qwen 2.5 7B model \cite{qwen2024technical}, 4 out of 100 randomly sampled responses under level 6 perturbation demonstrate number-copy errors. 
To enable the GPT-4o model to identify such errors, the model requires access to the numbers provided in the problem. Instead of providing the entire problem, we supply the model with a list of extracted numbers from the problem text. This approach is motivated by our observation that providing the full problem tends to lead the model to revise logically flawed answers into logically correct ones. This behavior aligns with the known tendency of LLMs to exhibit biases toward generating correct responses and their difficulty in intentionally producing incorrect answers \cite{tjuatja2023do, kumar2024investigating}. By limiting access to the original question, we minimize this undesired bias while enabling number-copy error correction. 



\subsection{Validation}

To validate 
our grading methodology, we perform a careful manual analysis. 
We collect 
responses generated by nine models across six perturbation levels from the experiments in Section \ref{sec:experiemnt}, along with the corresponding Python code produced by GPT-4o. We randomly sample 200 of these responses and
and manually classify each one. 
We found that our grading methodology correctly classified 197 of the responses correctly, 
achieving a high accuracy of 98.5\%. Furthermore, we assessed GPT-4o’s ability to correct number-copy errors when going from response to Python code.
We identify 50 responses across different models that contain such errors and evaluate whether the generated code correctly fixes them. Our manual review confirms that all 50 errors are successfully corrected. These evaluation results establish
the reliability of our methodology.

\section{Experiments and Results}

\label{sec:experiemnt}
Using GSM-Ranges and our proposed grading methodology, we evaluate the mathematical reasoning capabilities of nine distinct models, including both open-source and closed-source variants. 
Recall that we begin with 100 randomly selected GSM8K questions. For each question and for each of the six perturbation levels, we generate 50 random variations of the question.
This process yields a dataset of 5,000 problems per perturbation level. For each perturbation level and each model, we obtain responses to the 5,000 questions and classify the responses using our grading methodology. We then determine the percentages of correct answers, logical errors, and non-logical errors across the 5,000 responses. To compute confidence intervals, we leverage the structure of our dataset: each perturbation level consists of 50 distinct sets of questions derived from the original 100. We calculate the proportion of each error type within each set and then use these 50 sample proportions to estimate the corresponding confidence intervals, thereby quantifying the variability in model performance across perturbation instances. Additionally, we assess each model on the 100 unmodified base questions from GSM8K, providing a standard reference point for performance comparison. All inferences are done in the greedy decoding setting.

\subsection{Logical Errors}

\subsubsection{Rising Trend of Logical Errors}

Since the perturbations alter only the numerical values while keeping the question structure intact, the logical reasoning required to solve the problems remains unchanged across all perturbation levels. In principle, all perturbation levels should demand the same level of logical reasoning ability. 
Surprisingly, however, while the degree varies among the models, we observe a consistent upward trend in logical errors as the perturbation level increases, across all nine evaluated models (Figure \ref{fig:robust_exp}) except GPT-4o. To quantify this trend, we calculate the difference in logical error rates between level 6 (1M--10M) and level 1 (same digit) perturbations for each model (Figure \ref{fig:r6r1gap}). The most pronounced discrepancy is exhibited by the Gemma 2 2B model, which shows a 14\% absolute increase in logical error rate. Similarly, the WizardMath 7B v1.1 model demonstrates a substantial increase of 10\%. Even the relatively more robust models, such as Phi-3 Mini 4K and GPT-3.5 Turbo, still exhibit an increase of approximately 4\%, which remains a significant deviation. GPT-4o, one of the most advanced models at the time of our study, stands out as the only model with a near-zero gap. These results reveal the sensitivity of the models’ logical reasoning to numerical scales. We conjecture this phenomenon occurs because the models are mostly trained with lower-range numbers, and test problems with large numbers are out of distribution. (A qualitative analysis of additional logical errors induced by increasing numerical values is provided in Appendix \ref{subsec:error types}.)


 Another noteworthy observation is that the increase in logical errors becomes more gradual at higher perturbation levels. Across all nine models, the gap between level 3 and level 1 is generally larger than that between level 6 and level 3. This trend aligns with the cumulative frequency patterns observed in Figure \ref{cumulativegraph}, which shows that low-range values account for majority of numbers across the most widely used benchmark datasets. While the exact composition of training data for the models remains unknown, if math-problem training data is predominantly concentrated in the lower numerical ranges, it is plausible that beyond a certain threshold, further increases in numerical magnitude do not lead to a significant difference in model performance. Once numbers exceed this threshold, they may all be similarly unfamiliar to the model due to their low presence in the training data. Further investigation is needed to validate this hypothesis.

\subsubsection{Potential Data Contamination with GSM8K Dataset}

We also observe a notable logical error gap between level 1 perturbation and the original questions 
for many of the evaluated models (Figure \ref{fig:r6r1gap}). The Gemma 2 2B model exhibits the largest gap at 6\%, followed by Mistral 7B v0.1 with a 4\% discrepancy. However, this pattern is not consistent across all models. For instance, Qwen 2.5 7B and GPT-4o show a gap of only about 1\%, demonstrating better robustness. Moreover, Llama 3.2 3B and Phi-3 Mini 4K exhibit a -2\% gap, indicating an opposite trend.

This result points to possibility of data contamination to the GSM8K dataset in certain models. Notably, a similar finding was previously reported by \citet{mirzadeh2024-gsmsymbolic}, but our study explores this issue in more depth by providing an explicit definition of numerical-range similarity and establishing a clear distinction between logical and non-logical errors, further validating their conclusions.

\subsection{Arithmetic Errors}
\subsubsection{Rising Trend of Arithmetic Errors}

Previous studies have shown that LLMs exhibit a significant decline in arithmetic accuracy as numerical values grow \cite{qian2023-arithmeticerror, feng2024numerical}, and our result further confirms this trend.
As shown in Figure \ref{fig:robust_exp},
we also observe a consistent increase in non-logical errors. Given that number-copy errors account for only a small portion (Table \ref{tab:number-copy-errors}), the majority of these errors stem from arithmetic errors. 
Furthermore, because some responses classified as logical errors also include arithmetic errors, 
the true prevalence of arithmetic errors exceeds what is suggested in the figure. 

\subsubsection{Arithmetic Errors with Small Numbers}
\label{smallnumarithmetic}

Previous studies have found that state-of-the-art models have arithmetic accuracy on low-range numbers \cite{henighan2020scaling, yuan2023well,qian2023-arithmeticerror, feng2024numerical}. However, we find that some models still show non-trivial percentages of non-logical errors at level 1, such as Mistral 7B v0.1 at 9\% and WizardMath 7B v1.1 at 4\%. This motivates further analysis on the patterns of these errors, which is discussed in section \ref{sec:arithmetic error patterns}.



\begin{figure}[t]
  \includegraphics[width=\columnwidth]{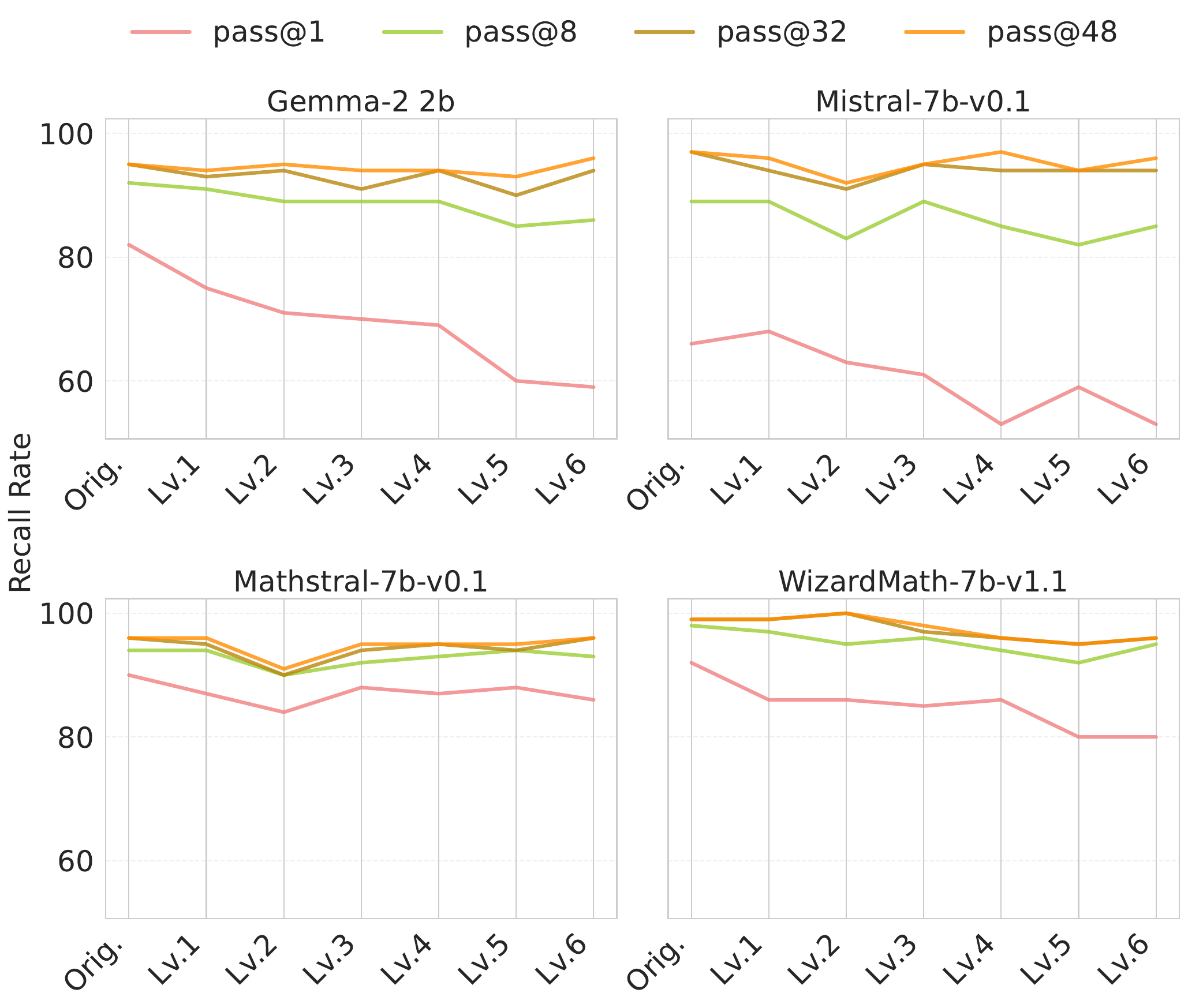}
  \caption{Recall rates across perturbation levels and original GSM8K questions for different sampling sizes (1, 8, 32, 48).}
  \label{fig:recall}
\end{figure}

\begin{figure}[t]
  \includegraphics[width=\columnwidth]{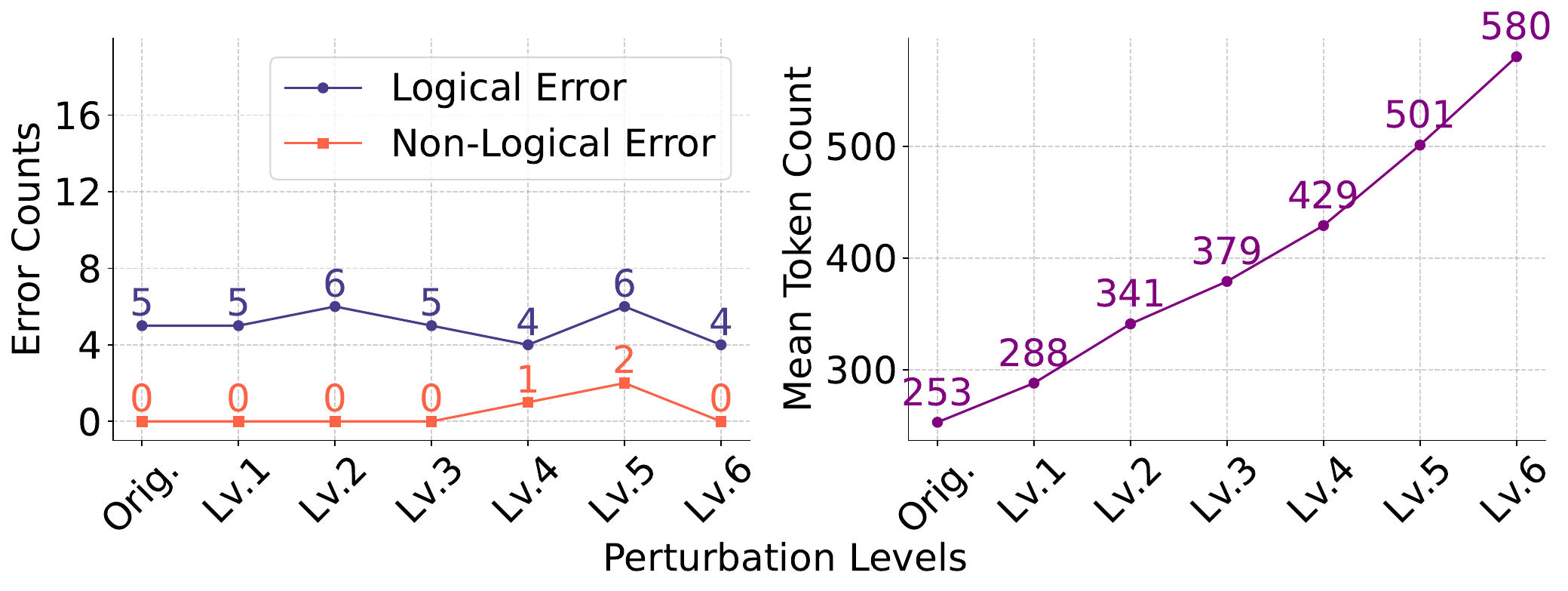}
  \caption{Results of o3-mini across perturbation levels. The left plot displays the logical and non-logical error counts, while the right plot shows the mean token counts per 100 responses at each perturbation level.}
    \label{fig:o3token}
\end{figure}

\section{In-depth Analysis}


\subsection{Is the Correct Logic Present in the LLM?}

We observe that larger numerical values in math questions increase the likelihood of logical errors. However, although a sampled response may have a logical error, the correct logic may nevertheless be present in the model's distribution. To investigate this issue,  we measure recall rates defined as follows: for each of 100 randomly generated questions, we obtain $n$ responses (i.e., perform $n$-passes) in a non-zero temperature setting  (temperature = 0.8, top\_p = 0.95) and, using our grading methodology, count the number of questions for which the correct logic appears in at least one of the passes. We perform this experiment over all six perturbation levels, and for a varying number of passes ranging from $n=1$ to $n=48$. 


As shown in Figure \ref{fig:recall}, for all four models, as the number of passes $n$ increases, we observe (1) a higher recall rate, and (2) smaller gaps across perturbation levels. At the highest sample size of 48, the gap between level 1 and 6 is no more than 2 for any model, indicating that the correct logic exists within the model’s distribution despite larger numerical values. This suggests that training on broader numerical ranges or leveraging test-time computation could improve numerical consistency.


\subsection{Performance of Reasoning Model}

The growing prominence of reasoning models \cite{openai_o1,guo2025deepseek} naturally raises the question of their performance across different perturbation levels. To explore this, we evaluate o3-mini—one of the most advanced reasoning models at the time of our study—on a set of 100 problems for each perturbation level. As shown in Figure \ref{fig:o3token}, o3-mini maintains consistently low logical and non-logical error rates across perturbation levels, demonstrating its robustness to varying numerical scales.



We also record the average token count for 100 responses at each perturbation level and observe that the model generates more tokens as the perturbation level increases (Figure \ref{fig:o3token}). While this increase may partly stem from larger numerical values requiring more tokens for representation and complex arithmetic, it also suggests that changes in numerical scale might lead the model to perceive the tasks as more challenging, possibly due to its training data being primarily focused on lower-range numbers. Additionally, the model produces more tokens for same-digit perturbations compared to the original GSM8K questions. This raises the possibility of data contamination, allowing the model to arrive at the correct final answer with less reasoning.


\subsection{Arithmetic Error Patterns}


\begin{table}[t]
  \centering  
  \resizebox{\columnwidth}{!}{ 
  \begin{tabular}{lll}  
    \hline  
    \textbf{Model}           & \textbf{Level 1} & \textbf{Level 2} \\  
    \hline  
    Gemma 2 2B       & 15/134 (11.2\%)          & 126/735 (17.1\%)   \\  
    WizardMath 7B v1.1       & 41/117 (35.0\%)          & 186/806 (23.1\%) \\  
    Mistral 7B v0.1 & 73/299 (21.4\%)  & 274/1313 (20.9\%)    \\  
    Mathstral 7B v0.1    & 31/77 (40.2\%)        & 126/509 (24.8\%)   \\  
    Llama 3.2 3B       & 3/72 (4.2\%)          & 127/1480 (8.6\%)        \\  
    Qwen 2.5 7B & 7/18 (38.9\%)  & 52/155 (33.5\%)    \\  
    Phi 3 Mini 4K & 9/22 (40.9\%)  & 89/281 (31.7\%)    \\  
    GPT-3.5 Turbo & 10/18 (55.6\%) & 92/224 (41.1\%) \\  
    GPT-4o & 1/3 (33.3\%) & 20/40 (50\%) \\  
    \hline  
  \end{tabular}  
  }
  \caption{  
  Results of a standalone arithmetic assessment on arithmetic errors made by models under Level 1 and Level 2 perturbations.
  } \label{tab:standalone}
\end{table}

\label{sec:arithmetic error patterns}




Previous studies have evaluated the arithmetic accuracy of LLMs in a standalone setting, i.e., directly posing arithmetic questions like "What is 1 + 2?" \cite{yang2023gpt, maltoni2024arithmetic, yuan2023well, qian2023-arithmeticerror, feng2024numerical}. However, little attention has been paid to whether their arithmetic performance remains robust when these operations are embedded within natural language responses. To investigate this, we conduct an experiment by collecting all responses containing arithmetic errors from all models under level 1 and 2 perturbations, and then extracting the specific arithmetic operations that were answered incorrectly. We subsequently prompt the models to solve these arithmetic operations in a standalone setting. 

As shown in Table \ref{tab:standalone}, while the extent varies, the models perform significantly better when the arithmetic task is isolated. We hypothesize that this phenomenon occurs because LLMs predominantly rely on memorization for arithmetic operations, since they train largely on standalone arithmetic data \cite{yuan2023well,yang2023gpt,maltoni2024arithmetic}. This results in degraded performance when these operations are integrated into a natural language context, which is out-of-distribution for the LLMs. 

\section{Conclusion}
In this work, we introduce GSM-Ranges, a benchmark designed to evaluate LLMs' reasoning abilities across diverse numerical scales. Additionally, we propose a novel grading methodology that classifies erroneous into logical and non-logical categories. Through extensive experiments on various models using GSM-Ranges and our grading framework, we find that logical accuracy tend to degrade significantly as perturbation level rises, revealing LLMs' sensitivity to numerical scales. Furthermore, while LLMs perform well on isolated arithmetic tasks, their accuracy declines significantly when calculations are integrated into natural language contexts. This study provides a more precise assessment of LLMs’ mathematical reasoning and paves the way for future research on improving mathematical reasoning capabilities and developing models that can generalize more effectively across diverse mathematical problem settings.


\section{Limitations}

Due to resource constraints, our study primarily focuses on small, lightweight models. While we have evaluated GPT-4o and o3-mini, future work could extend the analysis to other advanced models. Additionally, our perturbation study is conducted on the GSM8K dataset, and exploring the impact of varying numerical ranges on performance in more complex mathematical tasks would further enrich the findings. Lastly, while our grading methodology distinguishes between logical and non-logical errors, a more granular grading methodology could offer deeper insights into model performance and refinement.

\section*{Acknowledgments}

We express our gratitude to Yik-Cheung (Wilson) Tam, Professor of Practice in Computer Science at New York University Shanghai, for his valuable advice in shaping the ideas for this study.

\bibliography{custom}

\newpage 

\onecolumn

\appendix

\section{Appendix}
\label{sec:appendix}

\subsection{Experiment Results}
\subsubsection{Logical Error Rates}

\begin{table}[h]
    \centering
    \begin{tabular}{l c c c c c c c}
        \hline
        \multirow{2}{*}{Model} & \multirow{2}{*}{Baseline} & \multicolumn{6}{c}{Perturbation Levels} \\
        & & Lv.1 & Lv.2 & Lv.3 & Lv.4 & Lv.5 & Lv.6 \\
        \hline
        Gemma 2 2B       & 18 & 24.3(0.8) & 30.9(0.6) & 32.3(0.7) & 35.1(0.6) & 36.8(0.8) & 38.5(0.8) \\
        GPT-3.5 Turbo   & 11 & 13.5(0.5) & 15.0(0.8) & 17.5(0.6) & 17.3(0.7) & 16.9(0.6) & 17.3(0.6) \\
        GPT-4o          & 4  & 5.1(0.4)  & 6.9(0.3)  & 6.9(0.3)  & 6.2(0.3)  & 5.0(0.3)  & 5.3(0.3)  \\
        Llama 3.2 3B    & 17 & 14.5(0.5) & 17.0(0.6) & 19.3(0.7) & 18.7(0.7) & 19.4(0.6) & 19.4(0.6) \\
        Mathtral 7B v0.1 & 7  & 9.0(0.5)  & 11.6(0.5) & 12.4(0.5) & 13.9(0.5) & 15.2(0.6) & 14.8(0.6) \\
        Mistral 7B v0.1  & 29 & 33.3(0.9) & 37.5(0.9) & 38.5(0.9) & 40.7(0.7) & 42.4(0.8) & 43.3(0.8) \\
        Phi 3 Mini 4K   & 10 & 7.7(0.4)  & 9.1(0.4)  & 10.7(0.4) & 11.2(0.5) & 11.2(0.5) & 12.2(0.5) \\
        Qwen 2.5 7B     & 4  & 5.0(0.4)  & 7.8(0.5)  & 9.0(0.5)  & 10.2(0.5) & 10.1(0.6) & 9.9(0.5)  \\
        Wizardmath 7B v1.1 & 7  & 8.1(0.5)  & 14.0(0.6) & 15.7(0.7) & 16.4(0.7) & 17.6(0.6) & 19.1(0.6) \\
        o3-mini & 5 & 5 & 6 & 5 & 4 & 6 & 4\\

        \hline
    \end{tabular}
    \caption{Logical error rates and confidence intervals across different GSM-Ranges perturbation levels.}
    \label{tab:logical}
\end{table}

\subsubsection{Non-Logical Error Rates}
\begin{table}[h]
    \centering
    \begin{tabular}{l c c c c c c c}
        \hline
        \multirow{2}{*}{Model} & \multirow{2}{*}{Baseline} & \multicolumn{6}{c}{Perturbation Levels} \\
        & & Lv.1 & Lv.2 & Lv.3 & Lv.4 & Lv.5 & Lv.6 \\

        \hline
        Gemma 2 2B & 3  & 3.6(0.4)   & 14.7(0.9)  & 21.6(0.8)  & 25.9(0.9)  & 29.6(1.1)  & 37.2(1.2)  \\
        GPT-3.5 Turbo & 0  & 0.5(0.2)   & 5.1(0.5)   & 12.7(0.8)  & 18.1(0.8)  & 35.0(1.0)  & 38.6(1.0)  \\
        GPT-4o & 0  & 0.1(0.1)   & 0.8(0.2)   & 2.7(0.3)   & 3.6(0.3)   & 5.2(0.4)   & 5.2(0.6)   \\
        Llama 3.2 3B & 2  & 1.9(0.3)   & 25.1(1.1)  & 49.5(1.2)  & 59.1(1.2)  & 61.8(0.9)  & 68.8(0.9)  \\
        Mathtral 7B v0.1 & 2  & 2.0(0.4)   & 10.1(0.8)  & 14.8(0.8)  & 19.3(1.1)  & 23.0(0.9)  & 27.6(1.0)  \\
        Mistral 7B v0.1 & 12 & 9.3(0.5)   & 25.1(1.2)  & 31.0(1.1)  & 34.9(1.2)  & 38.6(1.2)  & 42.0(1.1)  \\
        Phi 3 Mini 4K & 1  & 0.5(0.2)   & 6.2(0.4)   & 10.5(0.7)  & 15.8(1.0)  & 21.4(0.9)  & 28.0(1.1)  \\
        Qwen 2.5 7B & 0  & 0.4(0.2)   & 3.8(0.5)   & 7.0(0.6)   & 9.9(0.7)   & 12.1(0.9)  & 16.4(0.9)  \\
        Wizardmath 7B v1.1 & 2  & 4.1(0.6)   & 15.5(0.7)  & 24.3(1.2)  & 31.0(1.0)  & 35.6(1.2)  & 42.5(1.3)  \\
        o3-mini & 0 & 0 & 0 & 0 & 1 & 2 & 0\\

        \hline
    \end{tabular}
    \caption{Non-logical error rates and \& confidence intervals across different GSM-Ranges perturbation levels.}
    \label{tab:nonlogical}
\end{table}

\newpage 
\subsubsection{Recall Rates for Correct Logics}
\begin{table}[h]
    \centering
    \begin{tabular}{l c c c c c c c c}
        \hline
        \multirow{2}{*}{Model} & \multirow{2}{*}{Sample Size} & \multicolumn{1}{c}{GSM8K} & \multicolumn{6}{c}{Perturbation Levels} \\
        & & \multicolumn{1}{c}{Baseline} & Lv.1 & Lv.2 & Lv.3 & Lv.4 & Lv.5 & Lv.6 \\
        \hline
        \multirow{4}{*}{Gemma 2 2B} 
        & 1  & 82 & 74 & 69 & 68 & 67 & 61 & 59\\
        & 8  & 92 & 89 & 87 & 87 & 87 & 84 & 84\\
        & 32 & 95 & 91 & 92 & 89 & 92 & 89 & 92\\
        & 48 & 95 & 92 & 93 & 92 & 92 & 92 & 94\\
        \hline
        \multirow{4}{*}{Mistral 7B v0.1} 
        & 1  & 66 & 66 & 61 & 60 & 51 & 59 & 50\\
        & 8  & 89 & 88 & 82 & 87 & 84 & 81 & 83\\
        & 32 & 97 & 93 & 90 & 93 & 93 & 93 & 93\\
        & 48 & 97 & 95 & 91 & 93 & 96 & 93 & 95\\
        \hline
        \multirow{4}{*}{Mathtral 7B v0.1} 
        & 1  & 90 & 85 & 82 & 86 & 85 & 86 & 84\\
        & 8  & 94 & 92 & 88 & 90 & 92 & 92 & 91\\
        & 32 & 96 & 93 & 88 & 92 & 94 & 92 & 94\\
        & 48 & 96 & 94 & 89 & 93 & 94 & 93 & 94\\
        \hline
        \multirow{4}{*}{Wizardmath 7B v1.1} 
        & 1  & 92 & 84 & 85 & 83 & 84 & 78 & 78\\
        & 8  & 98 & 95 & 93 & 94 & 92 & 90 & 94\\
        & 32 & 99 & 97 & 98 & 95 & 94 & 93 & 95\\
        & 48 & 99 & 97 & 98 & 96 & 94 & 93 & 95\\
        \hline
    \end{tabular}
    \caption{Recall rates across different sampling sizes and GSM-Ranges perturbation levels. We use }
    \label{tab:recall}
\end{table}


\subsubsection{Mean Token Counts of o3-mini Responses}

\begin{table}[h]
    \centering
    \begin{tabular}{lccccccc}
        \hline
        & Baseline & Level 1 & Level 2 & Level 3 & Level 4 & Level 5 & Level 6 \\
        \hline
        Mean Token Count & 252.8 & 287.6 & 340.6 & 378.8 & 429.3 & 501.0 & 579.8 \\
        \hline
    \end{tabular}
    \caption{Mean token counts across GSM-Ranges perturbation levels for o3-mini responses}
    \label{tab:mean_token_count}
\end{table}

\subsubsection{Number-Copy Error Analysis}
\label{subsec:number copy error}

\begin{table}[h!]
\centering
\begin{tabular}{lccc}
\hline
\textbf{Model} & \textbf{Lv. 4} & \textbf{Lv. 5} & \textbf{Lv. 6} \\
\hline
Qwen 2.5 7B      & 0 & 2 & 4 \\
Llama 3.2 3B      & 0 & 0 & 1 \\
Mathstral 7B v0.1  & 0 & 0 & 1 \\
Phi 3 Mini 4K      & 0 & 0 & 1 \\
Gemma 2 2B     & 0 & 0 & 0 \\
GPT-3.5 Turbo   & 0 & 0 & 0 \\
GPT-4o           & 0 & 0 & 0 \\
Mistral 7B v0.1    & 0 & 0 & 0 \\
Wizardmath 7B v1.1   & 0 & 0 & 0 \\

\hline
\end{tabular}
\caption{Occurrences
 of Number-Copy Errors in 100 Random Samples Across Levels 4, 5, and 6 for Each Model.}
\label{tab:number-copy-errors}
\end{table}
For each of the nine base models, we sampled 100 responses per level from the level 4, 5, and 6 perturbations to evaluate number-copy error rates. As shown in Table \ref{tab:number-copy-errors}, 4 out of the 8 base models exhibited number-copy errors under level 6 perturbation, while only one model showed errors under level 5, and none were observed at level 4.

\newpage
\subsection{Full Prompt for Inference}
The full prompt used for inferences in the experiments is shown below:
\begin{tcolorbox}[colback=blue!10, colframe=blue!50, title=Zero-shot Prompt for Inferences]
As an expert problem solver, solve the following mathematical question step by step. \\
Q: \texttt{\{Question\}} \\
A: Let’s think step by step.
\end{tcolorbox}

\newpage 

\newtcolorbox{promptbox}{
    colback=blue!5!white, 
    colframe=blue!80!black, 
    coltitle=black, 
    fonttitle=\bfseries, 
    sharp corners, 
    width=\textwidth, 
    boxrule=0.5mm, 
    left=5mm, 
    right=5mm, 
    top=2mm, 
    bottom=2mm, 
    breakable 
}

\subsection{Python Code Generation Prompt}
\label{subsec:4o-prompt}

Below is the prompt provided to the GPT-4o model for translating LLMs' responses into Python code. 
We introduce a step to verbalize the response logic prior to code generation, as this process is found to improve the alignment between the generated code and the original response. The temperature is set to 0 in the code generation process.

\begin{center}
\begin{tcolorbox}[colback=blue!10, colframe=blue!50, title=Python Code Generation Prompt, width=\textwidth]
You are tasked with writing Python code that replicates the logic described in a given response to a math problem.  
Your code must strictly follow the exact reasoning steps provided in the response, regardless of whether the logic is correct, inconsistent, or flawed.
\newline
\begin{enumerate}
    \item Do not fix or modify the reasoning described in the response, even if they seem incorrect or nonsensical.
    \item Develop a Python function named \texttt{solver()} that replicates the logic in the response exactly as described:
    \begin{itemize}
        \item Define and assign all necessary variables within the function.
        \item The function must not take any external arguments.
        \item The function must return the computed final numerical result.
    \end{itemize}
    \item Ensure that all arithmetic operations described in the response are explicitly written as code. Avoid directly copying the results of these operations or the final answer from the response.
    \item Refer to the list of numbers extracted from the question provided to ensure any copied numbers in the response match the original numbers.
    \begin{itemize}
        \item If a number in the response is incorrectly copied (e.g., misrepresenting 1333785 as 133785 or 13333785), correct the number in your code and document the correction as a comment in the code.
    \end{itemize}
    \item Include an explanation in the \texttt{explain} field that describes the steps and logic from the response, regardless of correctness.
    \item Provide the output in the following format:
    \begin{verbatim}
    {
        “extracted_answer”: “<final numerical value of the answer>”,
        “explain”: “<detailed explanation of the response logic>”,
        “python_code”: “```python\n<generated Python function>\n```”
    }
    \end{verbatim}
\end{enumerate}

\begin{itemize}
    \item This is the list of numbers extracted from the question: \texttt{\{number\_list\}}.
    \item This the response: \texttt{\{response\}}.
\end{itemize}

\end{tcolorbox}
\end{center}

\newpage
\subsection{Various Types of Additional Logical Errors in Level 6 Perturbation}
\label{subsec:error types}

We conducted a qualitative analysis of additional logical errors induced by increasing numerical values, focusing on cases with the greatest rise in errors across perturbation levels among the questions derived from the same base GSM8K questions for each model. Our findings reveal a diverse range of errors without a consistent pattern. Below, we present examples of logical errors exhibited under level 6 perturbation. While some errors were entirely absent at level 1 and emerged only at level 6, others, though rare at level 1, became significantly more frequent. The displayed LLM responses have been edited for readability, with any arithmetic errors corrected.

\begin{tcolorbox}[
    colback=blue!5, 
    colframe=blue!60!black, 
    title=Example 1: Missing Step Error, 
    sharp corners=south, 
    width=\textwidth
]

\textbf{Question:}  

Laurel's friend gave her \num{8852986} baby outfits that her child no longer needed. At her baby shower, Laurel received twice that amount in new baby outfits. Then, Laurel's mom gifted her with another \num{5309889} baby outfits. How many outfits does she have for her baby?

\vspace{0.5em}
\textbf{Ground Truth:}  
\begin{itemize}
    \item Laurel's friend gave her \num{8852986} outfits.
    \item At the baby shower, she received twice that amount:  
          \[
          2 \times 8852986 = 17705972
          \]
    \item Adding all outfits together:  
          \[
          8852986 + 17705972 + 5309889 = \mathbf{31868847}
          \]
\end{itemize}

\vspace{0.5em}
\textbf{LLM Response:}  
\begin{enumerate}
    \item Compute the baby shower outfits:  
          \[
          8852986 \times 2 = 17705972
          \]
    \item Compute the total outfits:  
          \[
          17705972 + 5309889 = \mathbf{23015861}
          \]
\end{enumerate}

\vspace{0.5em}
\textbf{Error:} 

The model ignores the original \num{8852986} outfits from Laurel's friend.

\end{tcolorbox}

\newpage

\begin{center}
\begin{tcolorbox}[
    colback=blue!5, 
    colframe=blue!60!black, 
    title=Example 2: Operator Error, 
    sharp corners=south, 
    width=\textwidth
]

\textbf{Question:}  

Finn watches \num{9360266} tadpoles swimming in the pond. Suddenly, Finn sees \num{7180820} of them come out of hiding from under a lily pad, then sees \num{12947038} of them hide under a rock. How many tadpoles can Finn see in the pond now?

\vspace{0.5em}
\textbf{Ground Truth:}  
\begin{itemize}
    \item Initially, Finn sees \num{9360266} tadpoles.
    \item \num{7180820} tadpoles come out from hiding, so the new visible count is:  
          \[
          9360266 + 7180820 = 16541086
          \]
    \item Then, \num{12947038} tadpoles hide under a rock, reducing the visible count:  
          \[
          16541086 - 12947038 = \mathbf{3594048}
          \]
\end{itemize}

\vspace{0.5em}
\textbf{LLM Response:}  
\begin{enumerate}
    \item Compute the new visible tadpoles after some emerge:  
          \[
          9360266 + 7180820 = 16541186
          \]
    \item Since 12947038 more hide under the rock, add them as well:  
          \[
          16541186 + 12947038 = \mathbf{29488224}
          \]
\end{enumerate}

\vspace{0.5em}
\textbf{Error:}  

The model makes an operator error by incorrectly adding the \num{12947038} tadpoles that hid under the rock instead of subtracting them.

\end{tcolorbox}
\end{center}

\begin{center}
\begin{tcolorbox}[
    colback=blue!5, 
    colframe=blue!60!black, 
    title=Example 3: Contextual Value Error, 
    sharp corners=south, 
    width=\textwidth
]

\textbf{Question:}  

Gloria is shoe shopping when she comes across a pair of boots that fit her shoe budget. However, she has to choose between the boots and two pairs of high heels that together cost \num{4528570} dollars less than the boots. If one pair of heels costs \num{3392343} dollars and the other costs twice as much, how many dollars are the boots?

\vspace{0.5em}
\textbf{Ground Truth:}  
\begin{itemize}
    \item The second pair of heels costs:
          \[
          3392343 \times 2 = 6784686
          \]
    \item The total cost of both pairs of heels:
          \[
          6784686 + 3392343 = 10177029
          \]
    \item The boots cost \num{4528570} dollars more than both pairs of heels:
          \[
          10177029 + 4528570 = \mathbf{14705599}
          \]
\end{itemize}

\vspace{0.5em}
\textbf{LLM Response:}  
\begin{enumerate}
    \item Compute the second pair of heels' cost:
          \[
          3392343 \times 2 = 6784686
          \]
    \item Compute the total cost of both pairs of heels:
          \[
          6784686 + 3392343 = 10177029
          \]
    \item The total cost of the heels is \$10177029 less than the boots, so the boots' price is:
          \[
          10177029 + 10177029 = \mathbf{20354058}
          \]
\end{enumerate}

\vspace{0.5em}
\textbf{Error:}  

The model makes a contextual value error by mismatching the numerical value with the corresponding entity. Instead of using \num{4528570} dollars as the price difference between the boots and the two pairs of high heels, it incorrectly uses the total cost of the heels.

\end{tcolorbox}
\end{center}

\begin{center}
\begin{tcolorbox}[
    colback=blue!5, 
    colframe=blue!60!black, 
    title=Example 4: Question Misinterpretation Error, 
    sharp corners=south, 
    width=\textwidth
]

\textbf{Question:}  

Mary is \num{1922674} years younger than Joan, who is \num{2112084} years older than Jessa. If Jessa is \num{1840103} years old, what is the sum of the ages of the three girls?

\vspace{0.5em}
\textbf{Ground Truth:}  
\begin{itemize}
    \item Joan's age:
          \[
          1840103 + 2112084 = 3952187
          \]
    \item Mary's age:
          \[
          3952187 - 1922674 = 2029513
          \]
    \item Sum of all ages:
          \[
          1840103 + 3952187 + 2029513 = \mathbf{7821803}
          \]
\end{itemize}

\vspace{0.5em}
\textbf{LLM Response:}  
\begin{enumerate}
    \item Joan is \num{2112084} years old and Jessa is \num{1840103} years old.
    \item Mary is \num{1922674} years younger than Joan, which means:
          \[
          2112084 - 1922674 = 1919410
          \]
    \item Now we can find the sum of their ages:
          \[
          2112084 + 1840103 + 1919410 = 5861607
          \]
\end{enumerate}

\vspace{0.5em}
\textbf{Error:}  

The model misinterprets the question by incorrectly treating Joan's age difference from Jessa as her actual age. Instead of computing Joan's age based on her relationship to Jessa, it directly assigned \num{2112084} as Joan’s age.

\end{tcolorbox}
\end{center}

\begin{center}
\begin{tcolorbox}[
    colback=blue!5, 
    colframe=blue!60!black, 
    title=Example 5: Omitted Information Error, 
    sharp corners=south, 
    width=\textwidth
]

\textbf{Question:}  

Mary is \num{1922674} years younger than Joan, who is \num{2112084} years older than Jessa. If Jessa is \num{1840103} years old, what is the sum of the ages of the three girls?

\vspace{0.5em}
\textbf{Ground Truth:}  
\begin{itemize}
    \item Joan's age:
          \[
          1840103 + 2112084 = 3952187
          \]
    \item Mary's age:
          \[
          3952187 - 1922674 = 2029513
          \]
    \item Sum of all ages:
          \[
          1840103 + 3952187 + 2029513 = \mathbf{7821803}
          \]
\end{itemize}

\vspace{0.5em}
\textbf{LLM Response:}  
\begin{enumerate}
    \item Mary is \num{1922674} years younger than Joan, who is \num{2112084} years older than Jessa.
    \item This can be written as:  
          \[
          \text{Joan} = \text{Jessa} + 2112084
          \]
          \[
          \text{Mary} = \text{Joan} - 1922674 = \text{Jessa} + 2112084 - 1922674 = \text{Jessa} + 189410
          \]

    \item Substituting the expressions for Joan and Mary, we get:  
          \[
          \text{Mary} + \text{Joan} + \text{Jessa} = (\text{Jessa} + 2112084) + (\text{Jessa} + 189410) + \text{Jessa}
          \]
    \item Simplifying this expression, we get:  
          \[
          \text{Mary} + \text{Joan} + \text{Jessa} = 3 \times \text{Jessa} + 2301494
          \]
\end{enumerate}

\vspace{0.5em}
\textbf{Error:} 

The model makes an omitted information error by failing to utilize the given value of Jessa’s age (\num{1840103}) in the final computation. Instead of calculating the actual sum of their ages, it leaves the expression in terms of Jessa’s age without substitution, leading to an incomplete and incorrect result.

\end{tcolorbox}
\end{center}

\twocolumn

\end{document}